\documentclass[runningheads]{llncs}
\usepackage{booktabs}
\usepackage{multirow}
\usepackage[table]{xcolor}
\usepackage[T1]{fontenc}
\usepackage{amsmath}
\usepackage{amssymb}
\usepackage{caption}
\usepackage{graphicx,verbatim}
\begin{document}
\title{SCKAN: Structural Consensus-based KAN Prototype Learning for Semi-Supervised Pancreas Segmentation}

\titlerunning{SCKAN}
%
\author{Yuqi Liu\inst{1} \and
Yufei Chen\inst{1}\thanks{Corresponding author} \and
Wei	Fu\inst{1} \and
Xiaodong Yue\inst{2} \and
Shuo Li\inst{3}}
%
\authorrunning{Yuqi Liu et al.}
\institute{School of Computer Science and Technology, Tongji University, Shanghai, China \email{yufeichen@tongji.edu.cn} \and
Artificial Intelligence Institute, Shanghai University, Shanghai, China\and
Department of Computer and Data Science and Department of Biomedical Engineering, Case Western Reserve University, Cleveland, USA}



  
\maketitle              
\begin{abstract}
Accurate pancreas segmentation is critical for early cancer diagnosis, where annotation scarcity necessitates Semi-Supervised Learning (SSL).
However, due to significant inter-sample morphological variability, existing SSL methods face severe generalizability limitations under sparse supervision, leading to the Supervision Bias problem.
To address this, we propose \textbf{S}tructural \textbf{C}onsensus-based \textbf{KAN} Prototype Learning (\textbf{SCKAN}), which constructs the first cross-sample structural consensus learning with Kolmogorov-Arnold Networks (KANs), to achieve more generalizable and accurate segmentation.
Specifically, SCKAN contains two key designs: 
Structure-constrained Prototype Consistency Learning (SPCL), which prompts unbiased structural representation by enforcing cross-sample consistency via prototype-level contrastive optimization, and Consensus-based Kolmogorov-Arnold Fusion (CKaF), which reduces morphology-specific bias by aggregating stable consensus and filtering sample-wise noise via KAN's adaptive B-spline nonlinearity.
Extensive experiments on two public pancreas datasets demonstrate the effectiveness of SCKAN. Code is at https://github.com/rhodaliu17/SCKAN.


\end{abstract}

\section{Introduction}

\begin{figure}[t]
    \centering
    \includegraphics[width=0.827\linewidth]{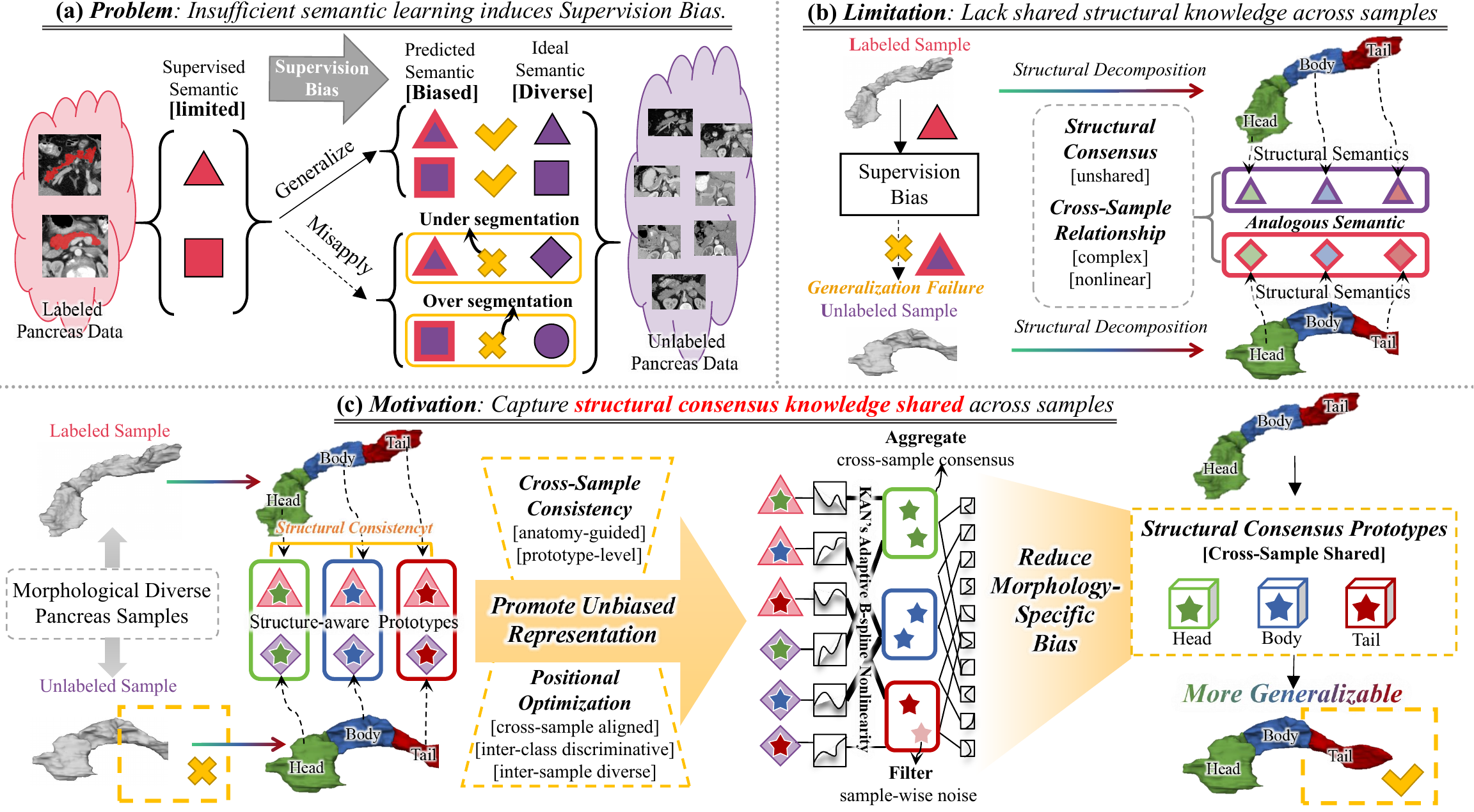}
    \caption{(a) Limited supervision signals on morphologically diverse pancreas data lead to Supervision Bias. (b) Existing methods suffer from poor generalizability, due to the lack of shared structural knowledge across samples. (c) SCKAN proposes the first cross-sample structural consensus learning with KANs to capture shared knowledge, enhancing generalizability.
    }
    \label{fig:first}
\end{figure}

Accurate pancreas segmentation~\cite{review1,chen2022target,review2} is critical for early cancer diagnosis~\cite{pancreaticcancer1,pancreaticcancer2,dedl,fu2023evidence}, yet scarce annotations make Semi-Supervised Pancreas Segmentation (SSPS) a necessary solution~\cite{FoundationSemi,SGRNet,ADMT}.
However, due to significant inter-sample morphological variability in the pancreas, existing methods~\cite{UAMT,urpc,DTC,TTMC,upcol,CPCL,mper,bapc,LIUpro} suffer from severe generalizability limitations, leading to under- or over-segmentation. Consistency Regularization (CR)-based methods~\cite{UAMT,urpc,DTC,TTMC} enforce sample-wise consistency yet lack effective cross-sample interaction, leaving semantics constrained within supervision. Prototype Learning (PL)-based methods~\cite{upcol,CPCL,mper,bapc,LIUpro} capture compact class-wise representations to guide segmentation, but fail to generalize to morphologically diverse samples, with prototype quality constrained by limited supervision.

Insufficient semantic learning constrained by supervision signals induces the \textit{Supervision Bias} problem (Fig.~\ref{fig:first}(a)). As a result, models fail to generalize beyond labeled guidance, resulting in biased representations. This limitation fundamentally lies in the lack of shared structural knowledge across samples in existing SSPS methods (Fig. \ref{fig:first}(b)). Anatomically, the pancreas can be decomposed into head, body, and tail subregions~\cite{panstructure1,panstructure2}, within each of which structural semantics remain analogous across diverse samples, forming the \textit{structural consensus}. However, due to complex and highly nonlinear cross-sample relationships, these structural semantics remain unshared. Kolmogorov-Arnold Networks (KANs)~\cite{KAN}, with their superior nonlinear fitting capability, enable effective modeling of such complex cross-sample relationships. Moreover, their adaptive B-spline functions with local support filter morphology-specific noise within local intervals while preserving stable consensus shared across high-variability samples, providing a strong basis for alleviating Supervision Bias.

To address this, we propose \textbf{S}tructural \textbf{C}onsensus-based \textbf{KAN} Prototype Learning (\textbf{SCKAN}), the first cross-sample consensus learning with KANs in SSPS (Fig.~\ref{fig:first}(c)). It captures structural consensus knowledge shared across samples through two key designs: 
(1) Structure-constrained Prototype Consistency Learning (SPCL), which employs Structure-aware Spatial Decomposition (SSD) to generate anatomy-guided subregion prototypes, establishing cross-sample structural correspondence, and Positional Consistency Calibration (PCC) to adopt position-weighted contrastive optimization and positional decorrelation regularization, reinforcing cross-sample consistency at the prototype level—thereby promoting unbiased structural representations.
(2) Consensus-based Kolmogorov-Arnold Fusion (CKaF), which leverages KAN's adaptive B-spline nonlinearity to filter sample-wise noise and aggregate stable structural consensus across high-variability samples, thereby reducing morphology-specific bias. \textbf{Our contributions are:}
\begin{itemize}
        \item We first propose structural consensus-based KAN to address supervision bias in SSPS under pancreas morphological diversity.
        \item Our novel SPCL introduces an anatomy-guided prototype-level consistency mechanism, promoting unbiased structural representation.
        \item Our novel CKaF introduces a KAN-enhanced cross-sample fusion strategy, aggregating stable consensus and reducing morphology-specific bias.
\end{itemize}

\section{Method}
\label{SCKAN}
SCKAN (Fig. \ref{fig:second}) constructs cross-sample structural consensus learning with KANs via two key designs: SPCL (Sec. \ref{sec:ssa}) promotes unbiased structural representations via prototype-level cross-sample consistency, and CKaF (Sec. \ref{sec:ckaf}) reduces morphology-specific bias via KAN-enhanced consensus aggregation. They are jointly optimized via Consensus-Guided Synergetic Learning (Sec. \ref{sec:all}).

\textbf{Preliminaries.} SSPS trains on labeled dataset $\mathcal{D}_l = \{(X_i^l, Y_i^l)\}_{i=1}^{N_l}$ and unlabeled dataset $\mathcal{D}_u = \{X_j^u\}_{j=1}^{N_u}$ where $N_u \gg N_l$. We adopt the Mean-Teacher framework (Fig. \ref{fig:second}(a)), which consists of a student model trained on both datasets, and a teacher model updated via exponential moving average (EMA) of the student's weights for stable predictions.

\subsection{SPCL for Promoting Unbiased Structural Representation}
\label{sec:ssa}
\begin{figure}[!t]
    \centering
    \includegraphics[width=0.92\linewidth]{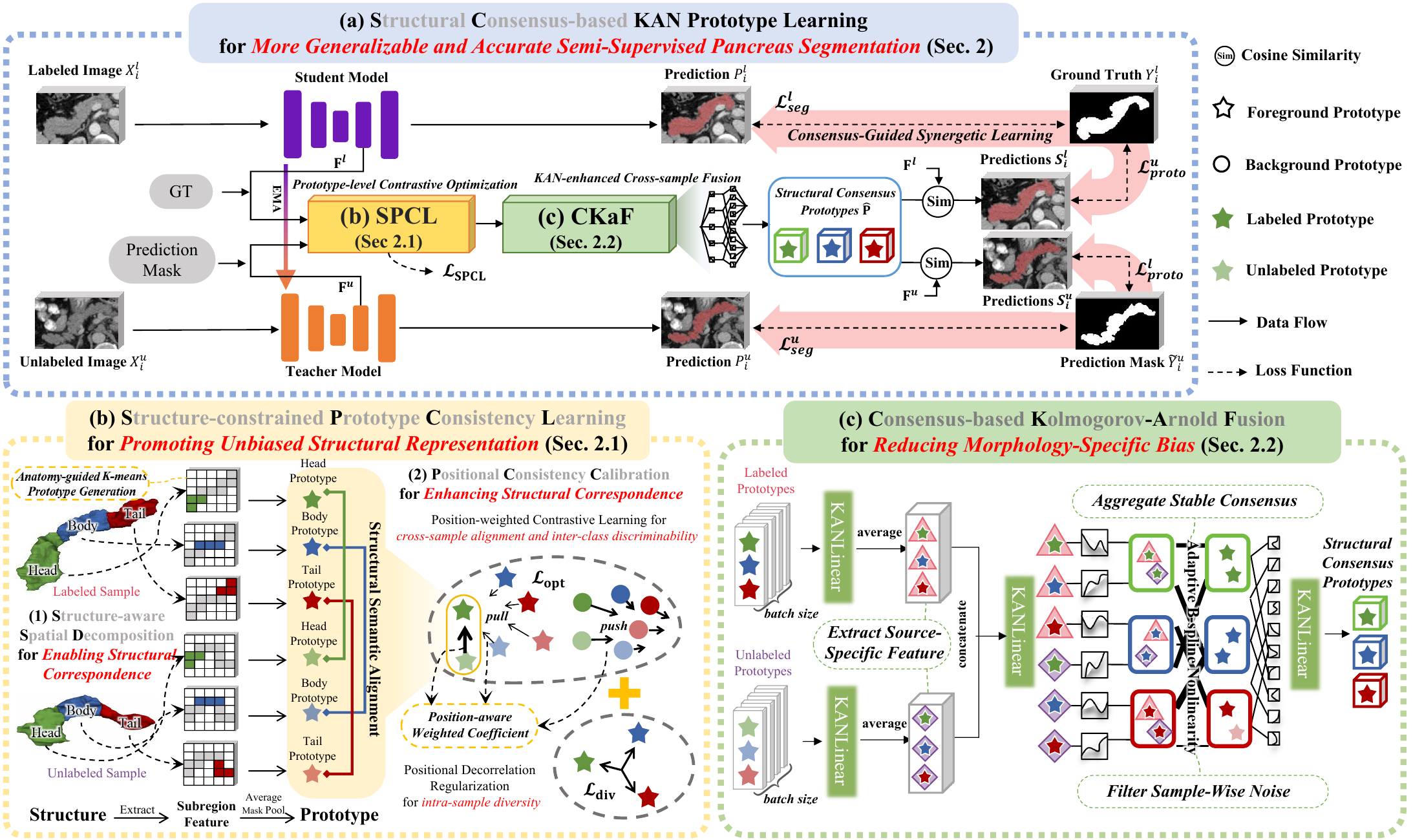}
    \caption{(a) SCKAN achieves generalizable and accurate SSPS results, consisting of (b) SPCL promoting unbiased structural representation and (c) CKaF reducing morphology-specific bias via aggregating consensus.}
    \label{fig:second}
\end{figure}

SPCL (Fig. \ref{fig:second}(b)) employs SSD to enable explicit anatomy-guided cross-sample structural correspondence, and further leverages PCC to reinforce position- and structure-aware cross-sample  alignment and intra-sample diversity.

\textbf{Structure-aware Spatial Decomposition (SSD)} effectively leverages anatomy-guided K-means clustering on spatial coordinates as a reliable anatomical proxy to decompose features into spatially-coherent ordered subregion prototypes (head, body, tail), grounded in the consistency of relative spatial positions of pancreatic subregions across samples.
Given the predicted mask $\mathbf{M}$ for class $c$, we decompose the masked region into $K$ spatially coherent subregions by applying K-means clustering on the valid spatial coordinates, yielding $K$ structure-aware regions $\{\mathbf{R}_k\}_{k=1}^{K}$.
To ensure consistent cross-sample correspondence, the regions are ordered by their cluster centers $\{\mathbf{A}_k\}_{k=1}^{K}$ according to a consistent spatial rule. Given feature map $\mathbf{F}$ from the 3rd decoder layer (student network $\mathbf{F}^l$ for labeled data, teacher network $\mathbf{F}^u$ for unlabeled data), we extract structure-aware prototypes by averaging features $\mathbf{P}_k = (\sum \mathbf{F} \cdot \mathbf{R}_k)/(\sum \mathbf{R}_k)$.

\textbf{Positional Consistency Calibration (PCC)} novelly calibrates hierarchical semantic relationships among subregion prototypes via position-aware optimization to compact subregion distributions in prototype feature space. It contains two components:

\emph{Position-weighted Contrastive Learning}  introduces position-aware weighting to differentiate subregion semantic relationships in contrastive learning, efficiently optimizing cross-sample subregion alignment and inter-class discriminability at the prototype level.
For each class $c$, let $\mathbf{P}^l$ and $\mathbf{P}^u$ denote the prototypes from labeled and unlabeled samples respectively. The complete prototype set is $\mathbf{P} = \{\mathbf{p}^l_1, ..., \mathbf{p}^l_K, \bar{\mathbf{p}}^l, \mathbf{p}^u_1, ..., \mathbf{p}^u_K, \bar{\mathbf{p}}^u\}$, where $K$ is the number of subregion prototypes and $\bar{\mathbf{p}}$ is the average of $K$ prototypes.
We introduce position-aware weighting that assigns different weight coefficients to prototype pairs of each sample:
\begin{equation}
w_{ij} = \begin{cases}
1.0 & \text{if same class and same region}\\
0.1 & \text{if same class but different region}\\
\end{cases}.
\end{equation}
The position-weighted contrastive optimization loss is formulated as:
\begin{equation}
\mathcal{L}_{\text{opt}} = -\frac{1}{|\mathbf{P}|} \sum_{i \in \mathbf{P}} \log \frac{\mathcal{P}^+_i}{\mathcal{P}^+_i + \mathcal{P}^-_i},
\end{equation}
where $\mathcal{P}^+_i$ and $\mathcal{P}^-_i$ are defined as $\mathcal{P}^+_i = \sum_{j \in \mathbf{P}^+_i} w_{ij} \cdot \exp(\text{sim}(\mathbf{P}_i, \mathbf{P}_j)/\tau)$ and $\mathcal{P}^-_i = \sum_{j \in \mathbf{P}^-_i} \exp(\text{sim}(\mathbf{P}_i, \mathbf{P}_j)/\tau)$.
$\mathcal{P}^+_i$ denotes positive pairs (same class as prototype $i$), $\mathcal{P}^-_i$ denotes negative pairs (different class), $\text{sim}(\cdot, \cdot)$ is cosine similarity, and $\tau$ is the temperature parameter.

\emph{Positional Decorrelation Regularization} imposes a position-sensitive similarity penalty on intra-sample structure-aware prototypes to ensure subregion diversity and distinct characteristics via:
\begin{equation}
    \mathcal{L}_{\text{div}} = \frac{1}{|\mathcal{C}|} \sum_{c \in \mathcal{C}} \frac{1}{K(K-1)} \sum_{k \neq k'} \max(0, \text{sim}(\mathbf{P}_{k}^c, \mathbf{P}_{k'}^c) - \alpha),
\end{equation}
where $\mathcal{C}$ is the total class number and the threshold $\alpha$ encourages moderate diversity.

With $\lambda_{\text{div}}$ balancing their contributions, the SPCL loss is formulated as:
\begin{equation}
\mathcal{L}_{\text{SPCL}} = (1-\lambda_{\text{div}}) \mathcal{L}_{\text{opt}} + \lambda_{\text{div}} \mathcal{L}_{\text{div}}.
\end{equation}
\textbf{Summary of Advantage: }SPCL novelly designs anatomy-guided K-means prototype generation in SSD, position-weighted contrastive optimization and positional similarity penalty in PCC, promoting unbiased structural representations.

\subsection{CKaF for Reducing Morphology-Specific Bias}
\label{sec:ckaf}
CKaF (Fig. \ref{fig:second}(c)) filters sample-wise noise and aggregates stable structural consensus across diverse samples through KAN-enhanced cross-sample fusion, leveraging adaptive B-spline functions with local support to reduce morphology-specific bias and greatly improve generalizability. 
Given structure-aware prototypes $\mathbf{P}^l$ and $\mathbf{P}^u$ from labeled and unlabeled samples, we apply separate KAN fusion to extract source-specific features:
\begin{equation}
    \mathbf{H}^{l} = \text{KANLinear}_l(\mathbf{P}^l), \quad 
    \mathbf{H}^{u} = \text{KANLinear}_u(\mathbf{P}^u),
\end{equation}
where $\text{KANLinear}(\mathbf{x}) = \sum_{i} h_i(x_i)$ is the KAN layer with learnable B-spline functions $h(\cdot)$~\cite{KAN}. Batch-level average pooling retains statistically stable structural knowledge across diverse samples, and the resulting representations are then integrated through two-stage KAN fusion to synthesize cross-sample structural consensus prototype $\hat{\mathbf{P}}$, where $\oplus$ is concatenation and $B$ is batch size:
\begin{equation}
    \mathbf{H} = \text{KANLinear}\!\left(\left[\frac{1}{B^l}\sum_{n=1}^{B^l}\mathbf{H}^{l}_n \oplus \frac{1}{B^u}\sum_{m=1}^{B^u}\mathbf{H}^{u}_m\right]\right), \quad \hat{\mathbf{P}} = \text{KANLinear}(\mathbf{H}).
\end{equation}
This hierarchical design utilizes KAN's superior fitting capacity to effectively model complex and nonlinear cross-sample relationships, efficiently aggregating cross-sample structural consensus knowledge.

\textbf{Summary of Advantage: }CKaF aggregates stable consensus by designing KAN-enhanced cross-sample fusion strategy for reducing morphology-specific bias.

\subsection{Consensus-Guided Synergetic Learning for Segmentation}
\label{sec:all}
Structural consensus knowledge is bridged between the prototype space and the segmentation network via consistency regularization, enabling the two branches to synergistically learn from each other (Fig. \ref{fig:second}(a)).
We define the hybrid loss as $\mathcal{L}_{\text{hybrid}} = \frac{1}{2}(\mathcal{L}_{\text{Dice}} + \mathcal{L}_{\text{CE}})$. For the segmentation network, we apply supervised loss on labeled data $\mathcal{L}_{\text{seg}}^l = \mathcal{L}_{\text{hybrid}}(P^l, Y^l)$ and consistency regularization on unlabeled data with pseudo-labels $\tilde{Y}^u$ from the teacher model $\mathcal{L}_{\text{seg}}^u = \mathcal{L}_{\text{hybrid}}(P^u, \tilde{Y}^u)$.
We then enforce consistency between segmentation network and prototype-based predictions $S = \arg\max_c \max_k \text{sim}(\mathbf{F}, \hat{\mathbf{P}}_k^c)$. We apply the same hybrid loss to the prototype-based predictions with $\mathcal{L}_{\text{proto}}^l = \mathcal{L}_{\text{hybrid}}(S^l, Y^l)$ and $\mathcal{L}_{\text{proto}}^u = \mathcal{L}_{\text{hybrid}}(S^u, \tilde{Y}^u)$.
The final training objective combines all components:
\begin{equation}
\mathcal{L}_{\text{total}} = \mathcal{L}_{\text{seg}}^l + \mathcal{L}_{\text{seg}}^u + \mathcal{L}_{\text{proto}}^l + \lambda_{\text{gs}}(\mathcal{L}_{\text{proto}}^u + \mathcal{L}_{\text{SPCL}}),
\end{equation}
where $\lambda_{\text{gs}} = e^{-5(1-s/s_{\max})^2}$ is a time-dependent Gaussian warming-up function~\cite{gaussianwarmup} that gradually increases consensus learning contribution during training.

\begin{table*}[t]
\centering
\caption{Comparison of different methods on NIH-PAN and MSD-PAN datasets demonstrating the advantages of our method. Results are reported as mean{\scriptsize$\pm$std}. Best results are in \textcolor{red}{red}, second-best in \textcolor{blue}{blue}.}
\label{tab:main_results}
\resizebox{1\textwidth}{!}{
\begin{tabular}{l|c|c|c|c|c|c||c|c|c|c|c}
\toprule
\multicolumn{12}{c}{\textbf{NIH-PAN Dataset}} \\
\midrule
\textbf{Method} & \textbf{Type} & \textbf{Lb/ULb} & \textbf{Dice $\uparrow$} & \textbf{Jaccard $\uparrow$} & \textbf{HD95 $\downarrow$} & \textbf{ASD $\downarrow$} & \textbf{Lb/ULb} & \textbf{Dice $\uparrow$} & \textbf{Jaccard $\uparrow$} & \textbf{HD95 $\downarrow$} & \textbf{ASD $\downarrow$} \\
\midrule
V-NET & SL & 62/0 (100\%) & 83.11{\scriptsize$\pm$5.57} & 71.46{\scriptsize$\pm$7.85} & 5.16{\scriptsize$\pm$3.76} & 1.05{\scriptsize$\pm$0.23} & -- & -- & -- & -- & -- \\
V-NET & SL & 3/0 (5\%) & 24.64{\scriptsize$\pm$21.18} & 15.84{\scriptsize$\pm$14.91} & 42.74{\scriptsize$\pm$42.43} & 7.00{\scriptsize$\pm$6.22} & 6/0 (10\%) & 53.76{\scriptsize$\pm$17.02} & 38.58{\scriptsize$\pm$15.79} & 20.35{\scriptsize$\pm$14.62} & 3.17{\scriptsize$\pm$5.20} \\
\midrule
UA-MT & SSL & \multirow{9}{*}{3/59 (5\%)} & 35.71{\scriptsize$\pm$12.21} & 22.41{\scriptsize$\pm$9.08} & 57.97{\scriptsize$\pm$13.81} & 25.12{\scriptsize$\pm$7.22} & \multirow{9}{*}{6/56 (10\%)} & 67.66{\scriptsize$\pm$14.73} & 52.78{\scriptsize$\pm$14.91} & 14.64{\scriptsize$\pm$11.90} & 2.84{\scriptsize$\pm$1.97} \\
SASSNet & SSL & & 50.87{\scriptsize$\pm$13.92} & 35.28{\scriptsize$\pm$12.65} & 30.23{\scriptsize$\pm$13.67} & 10.71{\scriptsize$\pm$6.26} & & 70.96{\scriptsize$\pm$15.44} & 56.82{\scriptsize$\pm$15.37} & 12.10{\scriptsize$\pm$1.15} & 4.69{\scriptsize$\pm$6.59} \\
URPC & SSL & & 49.68{\scriptsize$\pm$17.89} & 34.76{\scriptsize$\pm$14.55} & 26.44{\scriptsize$\pm$13.19} & 6.36{\scriptsize$\pm$3.77} & & 73.66{\scriptsize$\pm$11.95} & 59.45{\scriptsize$\pm$12.33} & 12.00{\scriptsize$\pm$8.08} & 2.34{\scriptsize$\pm$1.02} \\
BCP & SSL & & \textcolor{blue}{74.90{\scriptsize$\pm$7.39}} & \textcolor{blue}{60.39{\scriptsize$\pm$8.91}} & 11.38{\scriptsize$\pm$11.96} & 2.00{\scriptsize$\pm$1.04} & & \textcolor{blue}{82.59{\scriptsize$\pm$4.36}} & \textcolor{blue}{70.57{\scriptsize$\pm$6.10}} & 6.10{\scriptsize$\pm$4.05} & 1.73{\scriptsize$\pm$1.06} \\
AD-MT & SSL & & 67.46{\scriptsize$\pm$5.48} & 51.15{\scriptsize$\pm$6.07} & 13.70{\scriptsize$\pm$6.91} & 1.93{\scriptsize$\pm$0.60} & & 80.73{\scriptsize$\pm$6.52} & 68.18{\scriptsize$\pm$8.77} & 7.03{\scriptsize$\pm$6.66} & 1.39{\scriptsize$\pm$0.58} \\
CPCL & PL-SSL & & 50.88{\scriptsize$\pm$21.96} & 36.72{\scriptsize$\pm$17.84} & 33.16{\scriptsize$\pm$23.76} & 10.19{\scriptsize$\pm$14.51} & & 69.09{\scriptsize$\pm$13.89} & 54.23{\scriptsize$\pm$14.60} & 17.19{\scriptsize$\pm$15.45} & 3.69{\scriptsize$\pm$3.76} \\
UPCoL & PL-SSL & & 51.77{\scriptsize$\pm$18.84} & 36.86{\scriptsize$\pm$15.41} & 32.90{\scriptsize$\pm$24.42} & 3.32{\scriptsize$\pm$3.03} & & 72.78{\scriptsize$\pm$13.41} & 58.72{\scriptsize$\pm$14.45} & 14.00{\scriptsize$\pm$15.52} & 4.07{\scriptsize$\pm$3.87} \\
BaPC & PL-SSL & & 41.59{\scriptsize$\pm$18.19} & 27.88{\scriptsize$\pm$14.30} & 51.30{\scriptsize$\pm$30.85} & 3.86{\scriptsize$\pm$5.58} & & 75.10{\scriptsize$\pm$11.18} & 61.24{\scriptsize$\pm$12.50} & 12.26{\scriptsize$\pm$15.82} & 3.26{\scriptsize$\pm$4.49} \\
MPER & PL-SSL & & 74.88{\scriptsize$\pm$6.39} & 60.24{\scriptsize$\pm$7.67} & \textcolor{blue}{9.15{\scriptsize$\pm$3.84}} & \textcolor{blue}{2.34{\scriptsize$\pm$1.08}} & & 82.76{\scriptsize$\pm$4.88} & 70.87{\scriptsize$\pm$6.70} & \textcolor{blue}{5.32{\scriptsize$\pm$2.35}} & \textcolor{blue}{1.55{\scriptsize$\pm$0.71}} \\
SCKAN & PL-SSL & & \textcolor{red}{78.91{\scriptsize$\pm$4.17}} & \textcolor{red}{65.36{\scriptsize$\pm$5.62}} & \textcolor{red}{6.67{\scriptsize$\pm$2.90}} & \textcolor{red}{1.80{\scriptsize$\pm$0.66}} & & \textcolor{red}{83.91{\scriptsize$\pm$3.82}} & \textcolor{red}{72.47{\scriptsize$\pm$5.55}} & \textcolor{red}{4.61{\scriptsize$\pm$1.82}} & \textcolor{red}{1.33{\scriptsize$\pm$0.56}} \\
\midrule
\midrule
\multicolumn{12}{c}{\textbf{MSD-PAN Dataset}} \\
\midrule
\textbf{Method} & \textbf{Type} & \textbf{Lb/ULb} & \textbf{Dice $\uparrow$} & \textbf{Jaccard $\uparrow$} & \textbf{HD95 $\downarrow$} & \textbf{ASD $\downarrow$} & \textbf{Lb/ULb} & \textbf{Dice $\uparrow$} & \textbf{Jaccard $\uparrow$} & \textbf{HD95 $\downarrow$} & \textbf{ASD $\downarrow$} \\
\midrule
V-NET & SL & 168/0 (100\%) & 67.20{\scriptsize$\pm$19.79} & 53.40{\scriptsize$\pm$19.01} & 12.32{\scriptsize$\pm$17.36} & 2.31{\scriptsize$\pm$3.40} & -- & -- & -- & -- & -- \\
V-NET & SL & 17/0 (10\%) & 32.42{\scriptsize$\pm$30.18} & 23.61{\scriptsize$\pm$23.83} & 30.55{\scriptsize$\pm$32.77} & 6.34{\scriptsize$\pm$8.64} & 34/0 (20\%) & 56.51{\scriptsize$\pm$21.50} & 42.28{\scriptsize$\pm$19.55} & 20.29{\scriptsize$\pm$18.89} & 4.57{\scriptsize$\pm$4.56} \\
\midrule
UA-MT & SSL & \multirow{9}{*}{17/134 (10\%)} & 40.15{\scriptsize$\pm$26.01} & 28.55{\scriptsize$\pm$21.20} & 33.53{\scriptsize$\pm$16.67} & 12.72{\scriptsize$\pm$8.85} & \multirow{9}{*}{34/134 (20\%)} & 55.62{\scriptsize$\pm$23.37} & 41.88{\scriptsize$\pm$20.90} & 29.30{\scriptsize$\pm$31.19} & 2.25{\scriptsize$\pm$1.80} \\
SASSNet & SSL & & 39.56{\scriptsize$\pm$23.42} & 27.38{\scriptsize$\pm$18.80} & 35.00{\scriptsize$\pm$17.39} & 14.40{\scriptsize$\pm$9.03} & & 57.79{\scriptsize$\pm$19.55} & 43.13{\scriptsize$\pm$18.32} & 20.22{\scriptsize$\pm$17.60} & 4.61{\scriptsize$\pm$3.40} \\
URPC & SSL & & 43.02{\scriptsize$\pm$26.67} & 31.14{\scriptsize$\pm$22.23} & 32.20{\scriptsize$\pm$19.53} & 6.10{\scriptsize$\pm$5.06} & & 59.08{\scriptsize$\pm$19.54} & 44.52{\scriptsize$\pm$18.85} & 20.56{\scriptsize$\pm$17.32} & 4.18{\scriptsize$\pm$4.31} \\
BCP & SSL & & 57.63{\scriptsize$\pm$18.02} & 42.67{\scriptsize$\pm$17.48} & 21.58{\scriptsize$\pm$13.26} & 7.05{\scriptsize$\pm$4.45} & & \textcolor{blue}{62.63{\scriptsize$\pm$17.53}} & 47.77{\scriptsize$\pm$17.27} & \textcolor{blue}{19.72{\scriptsize$\pm$25.49}} & \textcolor{blue}{2.04{\scriptsize$\pm$1.44}} \\
AD-MT & SSL & & \textcolor{blue}{59.74{\scriptsize$\pm$21.58}} & \textcolor{blue}{45.57{\scriptsize$\pm$19.60}} & 20.86{\scriptsize$\pm$26.00} & \textcolor{red}{3.22{\scriptsize$\pm$4.33}} & & 62.14{\scriptsize$\pm$23.61} & \textcolor{blue}{48.67{\scriptsize$\pm$21.27}} & 22.96{\scriptsize$\pm$31.38} & 2.66{\scriptsize$\pm$4.16} \\
CPCL & PL-SSL & & 29.95{\scriptsize$\pm$30.39} & 21.86{\scriptsize$\pm$23.90} & 40.57{\scriptsize$\pm$40.89} & 6.19{\scriptsize$\pm$9.77} & & 50.64{\scriptsize$\pm$26.84} & 38.00{\scriptsize$\pm$23.01} & 36.38{\scriptsize$\pm$36.32} & 2.84{\scriptsize$\pm$3.84} \\
UPCoL & PL-SSL & & 43.35{\scriptsize$\pm$30.51} & 32.46{\scriptsize$\pm$24.78} & 41.42{\scriptsize$\pm$37.80} & 8.79{\scriptsize$\pm$11.55} & & 59.56{\scriptsize$\pm$21.21} & 45.34{\scriptsize$\pm$19.61} & 25.92{\scriptsize$\pm$29.53} & 2.78{\scriptsize$\pm$4.20} \\
BaPC & PL-SSL & & 42.93{\scriptsize$\pm$30.06} & 32.04{\scriptsize$\pm$24.83} & 44.00{\scriptsize$\pm$40.07} & 6.93{\scriptsize$\pm$11.43} & & 56.15{\scriptsize$\pm$25.33} & 42.85{\scriptsize$\pm$21.91} & 32.83{\scriptsize$\pm$34.81} & 3.30{\scriptsize$\pm$5.24} \\
MPER & PL-SSL & & 55.65{\scriptsize$\pm$18.99} & 40.90{\scriptsize$\pm$18.03} & \textcolor{blue}{19.97{\scriptsize$\pm$13.08}} & 6.71{\scriptsize$\pm$4.68} & & 59.43{\scriptsize$\pm$18.42} & 44.54{\scriptsize$\pm$17.37} & 23.27{\scriptsize$\pm$27.97} & 3.24{\scriptsize$\pm$2.60} \\
SCKAN & PL-SSL & & \textcolor{red}{61.71{\scriptsize$\pm$18.00}} & \textcolor{red}{46.94{\scriptsize$\pm$17.97}} & \textcolor{red}{19.35{\scriptsize$\pm$23.70}} & \textcolor{blue}{4.07{\scriptsize$\pm$4.05}} & & \textcolor{red}{63.72{\scriptsize$\pm$17.50}} & \textcolor{red}{48.97{\scriptsize$\pm$17.47}} & \textcolor{red}{17.66{\scriptsize$\pm$22.33}} & \textcolor{red}{2.03{\scriptsize$\pm$1.09}} \\
\bottomrule
\end{tabular}
}
\end{table*}
\section{Experiments}

\subsection{Experimental Setup}
\textbf{Datasets.} We evaluate on two pancreas datasets. \textbf{(a)} \textit{NIH-PAN}~\cite{PAN} contains 80 contrast-enhanced CT scans, with 62 for training and 18 for testing. \textbf{(b)} \textit{MSD-PAN}~\cite{MSD} contains 281 annotated cases, where pancreas and tumor labels are merged into a single foreground class following  ~\cite{levelset,leveldiffusion}, with 168 for training, 57 for validation, and 56 for testing. \textbf{Implementation Details.} SCKAN is implemented in PyTorch and trained on a single NVIDIA RTX 3090 GPU. V-Net serves as the backbone of all methods. We use a batch size of 8 with 4 labeled and 4 unlabeled data, SGD optimizer with momentum 0.9, initial learning rate 0.01, and EMA decay 0.99. Data augmentation includes random cropping and copy-paste \cite{BCP,cutmix}. Prototype number is set to $K=3$ to correspond to structural decomposition for head, body and tail. Images are resized to $96\times96\times96$ following~\cite{urpc,upcol}. Performance is measured via Dice, Jaccard, HD95, and ASD.

\subsection{Results and Analysis}

\begin{figure}[t]
    \centering
    \includegraphics[width=0.7\linewidth]{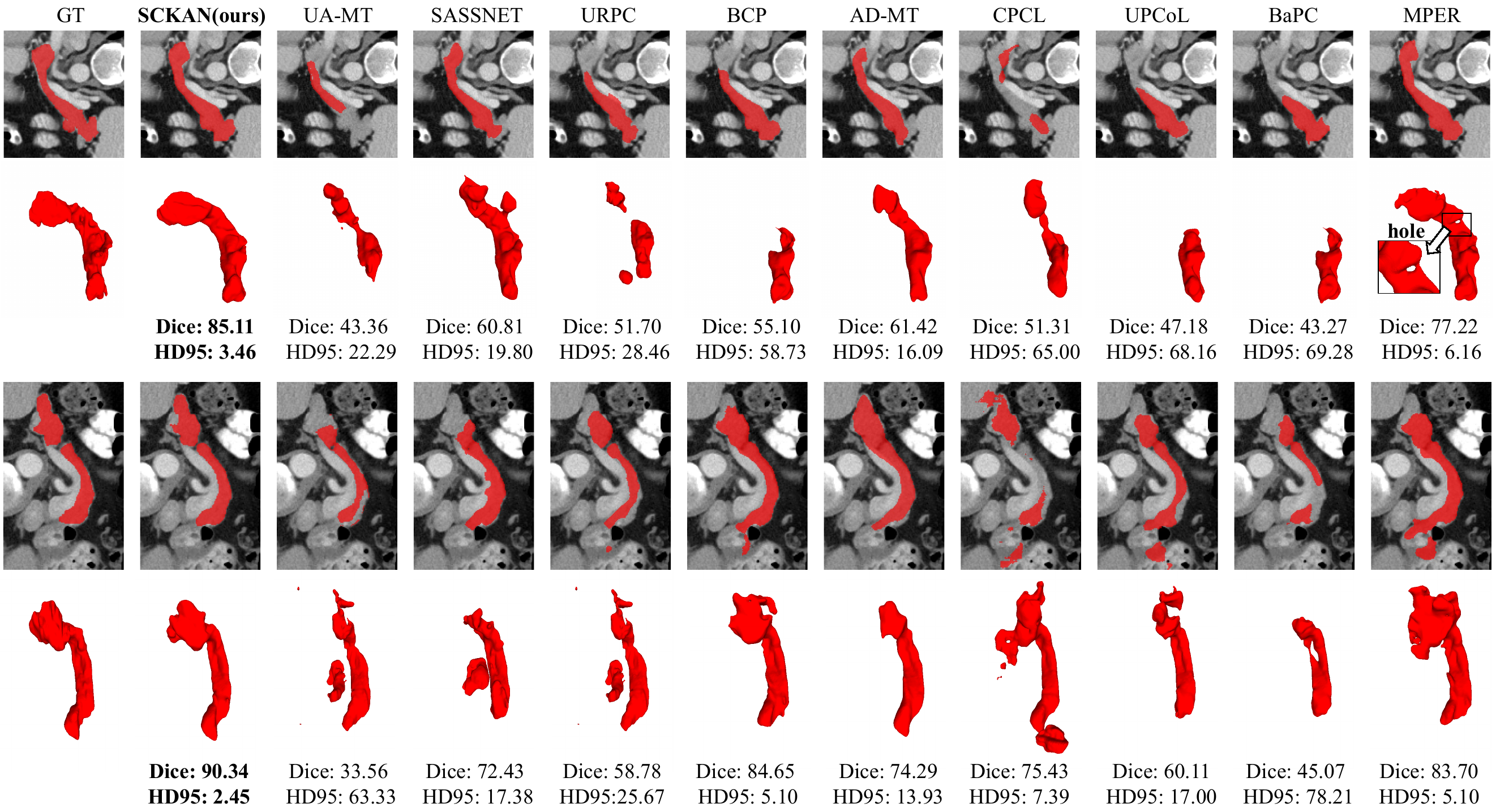}
    \caption{2D slice, 3D visualizations and metric performance on 5\% NIH-PAN dataset illustrating our method's capability in preserving structural completeness and achieving accurate segmentation without extra predictions.}
    \label{fig:fourth}
\end{figure}

\begin{figure}[]
    \centering
    \begin{minipage}[]{0.48\linewidth}
        \centering
        \caption{Comparison of methods (only Dice gap < 20\%)  between labeled and unlabeled data on 10\% NIH-PAN.}
        \label{fig:gap_analysis}
        \includegraphics[width=0.82\linewidth]{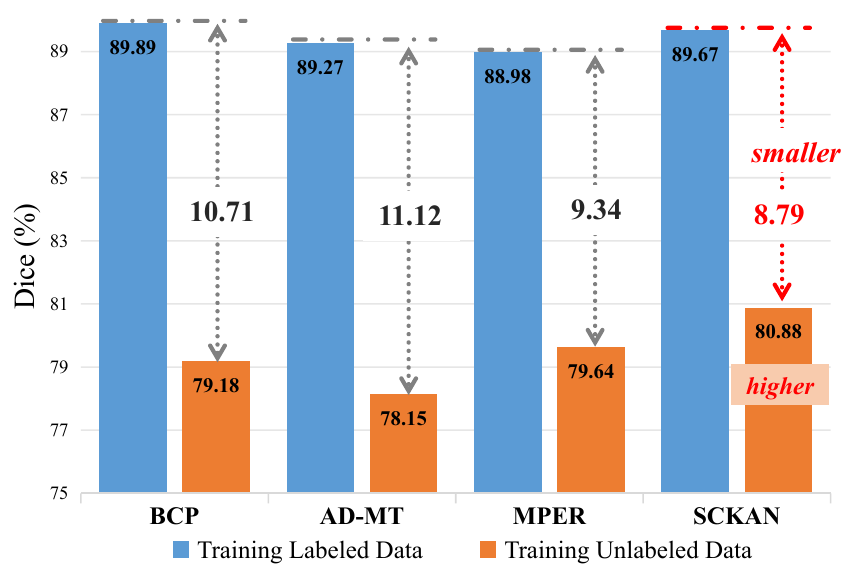}
    \end{minipage}
    \hfill
    \begin{minipage}[]{0.48\linewidth}
        \centering
        \caption{Ablation study on diversity loss coefficient $\lambda_{\text{div}}$ on both datasets. Best results at 0.5.}
        \label{fig:div_analysis}
        \includegraphics[width=0.82\linewidth]{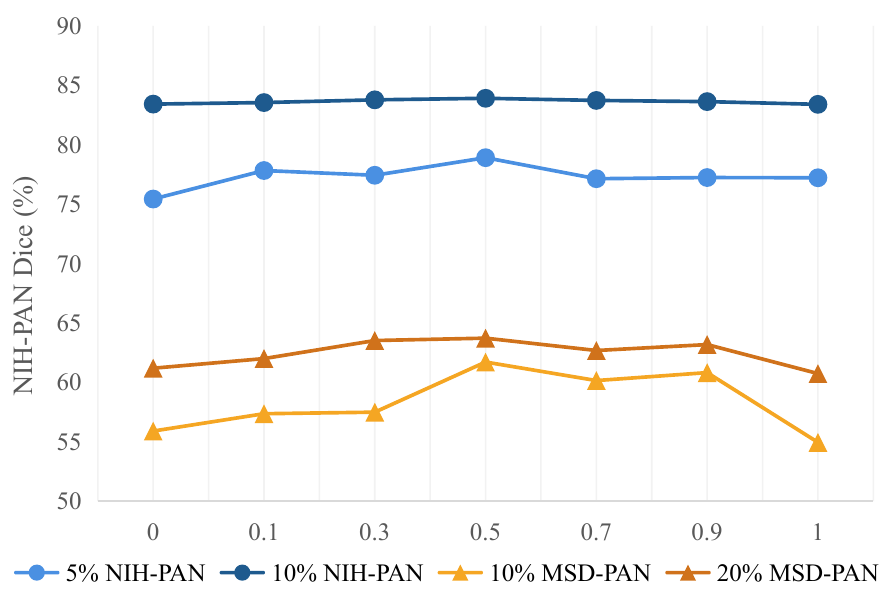}
    \end{minipage}
\end{figure}

\subsubsection{Comparison with State-of-the-Arts (Table~\ref{tab:main_results}, Fig.~\ref{fig:fourth}, Fig.~\ref{fig:gap_analysis}).}
Experiments are conducted on two pancreas datasets to demonstrate our advantage. Our method achieves superior metrics, particularly under extreme supervision scarcity (5\% NIH-PAN, only 3 samples), demonstrating enhanced structural completeness and accurate segmentation. On the more challenging MSD-PAN with lower image quality, our method still achieves competitive results with consistently low variance.  Visualizations in Fig.~\ref{fig:fourth} further validate our advantage with substantially higher Dice and lower HD95. Fig.~\ref{fig:gap_analysis} reveals that our method effectively alleviates Supervision Bias, yielding more generalizable results.

\begin{table}[t]
\centering
\caption{Ablation study on key components under 10\% NIH-PAN validating the effectiveness of our novel designs. Improvements are relative to MT+PL.}
\label{tab:ablation_components}
\setlength{\tabcolsep}{6pt}
\renewcommand{\arraystretch}{1}
\resizebox{0.8\columnwidth}{!}{
\begin{tabular}{cc|ccc|cccc}
\toprule
\multirow{2}{*}{\textbf{MT}} & \multirow{2}{*}{\textbf{PL}} & \multicolumn{2}{c}{\textbf{SPCL}} & \multirow{2}{*}{\textbf{CKaF}} & \multirow{2}{*}{\textbf{Dice$\uparrow$}} & \multirow{2}{*}{\textbf{Jaccard$\uparrow$}} & \multirow{2}{*}{\textbf{HD95$\downarrow$}} & \multirow{2}{*}{\textbf{ASD$\downarrow$}} \\
\cmidrule(lr){3-4}
& & \textbf{SSD} & \textbf{PCC} & & & & & \\
\midrule
$\checkmark$ & & & & & 66.60 & 52.55 & 18.97 & 6.12 \\
$\checkmark$ & $\checkmark$ & & & & 77.69 & 64.16 & 9.15 & 2.09 \\
\midrule
$\checkmark$ & $\checkmark$ & $\checkmark$ & & & 80.21 \textcolor{red}{($\uparrow$2.52)} & 67.36 \textcolor{red}{($\uparrow$3.20)} & 5.92 \textcolor{red}{($\downarrow$3.23)} & 1.62 \textcolor{red}{($\downarrow$0.47)} \\
$\checkmark$ & $\checkmark$ & $\checkmark$ & $\checkmark$ & & 82.61 \textcolor{red}{($\uparrow$4.92)} & 70.59 \textcolor{red}{($\uparrow$6.43)} & 5.42 \textcolor{red}{($\downarrow$3.73)} & 1.57 \textcolor{red}{($\downarrow$0.52)} \\
$\checkmark$ & $\checkmark$ & & & $\checkmark$ & 81.92 \textcolor{red}{($\uparrow$4.23)} & 69.63 \textcolor{red}{($\uparrow$5.47)} & 5.91 \textcolor{red}{($\downarrow$3.24)} & 1.68 \textcolor{red}{($\downarrow$0.41)} \\
\midrule
$\checkmark$ & $\checkmark$ & $\checkmark$ & $\checkmark$ & $\checkmark$ & \textbf{83.91} \textcolor{red}{\textbf{($\uparrow$6.22)}} & \textbf{72.47} \textcolor{red}{\textbf{($\uparrow$8.31)}} & \textbf{4.61} \textcolor{red}{\textbf{($\downarrow$4.54)}} & \textbf{1.33} \textcolor{red}{\textbf{($\downarrow$0.76)}} \\
\bottomrule
\end{tabular}}
\end{table}

\textbf{Ablation on Key Components (Table~\ref{tab:ablation_components}).} We validate each component on 10\% NIH-PAN. The baseline Mean Teacher (MT) with prototype learning (PL)  achieves 77.69\% Dice, confirming the benefit of prototype-enhanced consistency regularization. Introducing SSD further improves Dice to 80.21\% and SSD+PCC contributes an additional +2.4\% Dice, validating the advantage of prompting unbiased structural representations in SPCL. CKaF alone yields 81.92\%, confirming effective cross-sample consensus aggregation. The full SCKAN achieves optimal performance, with SPCL and CKaF synergistically contributing to accurate segmentation.

\textbf{Ablation on Diversity Loss Coefficient $\lambda_{\text{div}}$ (Fig. \ref{fig:div_analysis}).} 
The coefficient $\lambda_{\text{div}}$ balances $\mathcal{L}_{\text{opt}}$ and $\mathcal{L}_{\text{div}}$ in SPCL, controlling the trade-off between cross-sample structural correspondence and intra-sample diversity. Setting $\lambda_{\text{div}}=0$ causes prototypes to collapse into overly similar representations, while $\lambda_{\text{div}}=1.0$ sacrifices inter-sample alignment. Performance remains stable across intermediate values, with the best results achieved at $\lambda_{\text{div}}=0.5$, reaching optimal balance.

\begin{table}[t]
\centering
\caption{Ablation study on three prototype fusion strategies showing the effectiveness of KAN-enhanced cross-sample fusion in alleviating Supervision Bias.}
\label{tab:ablation_fusion}
\setlength{\tabcolsep}{6pt}
\renewcommand{\arraystretch}{1.1}
\resizebox{0.9\columnwidth}{!}{
\begin{tabular}{l|cccc||cccc}
\toprule
\multirow{2}{*}{\textbf{Fusion strategy}} & \multicolumn{4}{c||}{\textbf{5\% NIH-PAN}} & \multicolumn{4}{c}{\textbf{10\% MSD-PAN}} \\
\cmidrule(lr){2-5} \cmidrule(lr){6-9}
& \textbf{Dice$\uparrow$} & \textbf{L-Dice} & \textbf{U-Dice} & \textbf{L-U Dice$\downarrow$} & \textbf{Dice$\uparrow$} & \textbf{L-Dice} & \textbf{U-Dice} & \textbf{L-U Dice$\downarrow$} \\
\midrule
Average Fusion & 75.96 & 87.97 & 71.56 & 16.38 & 59.30 & 92.75 & 69.97 & 22.78 \\
MLP Fusion & 77.45 & 90.10 & 74.61 & 15.49 & 60.94 & 92.25 & 70.82 & 21.43 \\
\textbf{KAN Fusion (ours)} & \textbf{78.91} & 89.96 & 81.31 & \textbf{8.65}\textcolor{red}{\textbf{($\downarrow$6.84)}} & \textbf{61.71} & 89.79 & 70.88 & \textbf{18.91}\textcolor{red}{\textbf{($\downarrow$2.52)}} \\
\bottomrule
\end{tabular}}

{\fontsize{6}{7}\selectfont *`L' and `U' denote training labeled and unlabeled data, respectively. `L-U Dice' means their Dice gap.}
\end{table}

\textbf{Ablation on Fusion Strategies (Table \ref{tab:ablation_fusion}).} 
We compare three prototype fusion strategies to validate the effectiveness of our \textbf{KAN}-enhanced cross-sample \textbf{fusion }strategy in alleviating Supervision Bias. Our fusion achieves the best overall performance on both datasets. Compared with the competitive Multilayer Perceptron (MLP) fusion strategy, we further narrow the L-U Dice gap by 6.84\% and 2.52\% on two datasets, respectively. 
This demonstrates that our CKaF better reduces morphology-specific bias by filtering sample-wise noise and aggregating cross-sample consensus.

\section{Conclusion}
SCKAN addresses the Supervision Bias problem in SSPS under pancreatic diversity by constructing the first structural consensus-based KAN. Building on this, SPCL enforces cross-sample consistency via prototype-level contrastive optimization for unbiased structural representations, and CKaF aggregates stable cross-sample consensus via KAN's adaptive B-spline nonlinearity to reduce morphology-specific bias. Extensive experiments validate the effectiveness and generalizability of SCKAN. Code is available at https://github.com/rhodaliu17/\\SCKAN.

\begin{credits}
\subsubsection{\ackname} This work was supported by the National Natural Science Foundation of China (No. 62472315, No. 62476165).

\subsubsection{\discintname} The authors have no competing interests to declare that are relevant to the content of this article.
\end{credits}


\bibliographystyle{unsrt}

\bibliography{Paper-0226}
\end{document}